\documentclass[]{damo}

\usepackage{wrapfig}
\usepackage{tabularx}
\usepackage{textcomp}
\usepackage{stfloats}
\usepackage{url}
\usepackage{verbatim}
\usepackage{graphicx}
\usepackage{titlesec}
\usepackage{tocloft}
\usepackage{adjustbox}
\usepackage{multirow}
\usepackage{pifont}
\usepackage{tikz}
\usepackage{comment}
\usepackage{amsmath,amssymb} 
\usepackage{colortbl}  
\usepackage{color}
\usepackage{booktabs} 
\usepackage{hyperref}
\usepackage{graphicx}    
\usepackage{subcaption} 
\usepackage{booktabs}
\usepackage{amsmath} 
\usepackage{multirow} 
\usepackage{booktabs} 
\usepackage{subcaption} 
\usepackage{graphicx}   
\usepackage{wrapfig}    
\usepackage{caption}    

\RequirePackage{xspace}
\makeatletter
\DeclareRobustCommand\onedot{\futurelet\@let@token\@onedot}
\def\@onedot{\ifx\@let@token.\else.\null\fi\xspace}

\makeatother

\usepackage{makecell}

\definecolor{adptorange}{RGB}{248, 205, 172}
\definecolor{cmpblue}{RGB}{189, 215, 238}
\definecolor{cmpblue}{RGB}{189, 215, 238}

\definecolor{our_red}{RGB}{232,157,160}
\definecolor{our_blue}{RGB}{136,206,230}
\definecolor{our_orange}{RGB}{246,200,168}
\definecolor{our_green}{RGB}{178,211,164}

\definecolor{attn_code0}{RGB}{247,215,200}
\definecolor{attn_code1}{RGB}{238,169,139}
\definecolor{mlp_code0}{RGB}{204,201,221}
\definecolor{mlp_code1}{RGB}{102,95,153}

\definecolor{token_blue}{RGB}{84, 120, 140}

\definecolor{myMagenta}{rgb}{0.9,0,0.4}

\usepackage{pifont}       
\usepackage{bbding}       
\usepackage{fontawesome}
\usepackage{xspace}

\usepackage{float}

\newlength\savewidth

\newcolumntype{x}[1]{>{\centering\arraybackslash}p{#1pt}}
\newcolumntype{y}[1]{>{\raggedright\arraybackslash}p{#1pt}}
\newcolumntype{z}[1]{>{\raggedleft\arraybackslash}p{#1pt}}

\renewcommand{\paragraph}[1]{\vspace{1mm}\noindent\textbf{#1}}
\usepackage{colortbl}
\usepackage{xcolor}
\usepackage{wrapfig}

\renewcommand{\paragraph}[1]{\vspace{1.25mm}\noindent\textbf{#1}}

\usepackage{algorithm}
\usepackage{listings}

\definecolor{codeblue}{rgb}{0.25, 0.5, 0.5}
\definecolor{codekw}{rgb}{0.35, 0.35, 0.75}
\lstdefinestyle{Pytorch}{
    language = Python,
    backgroundcolor = \color{white},
    basicstyle = \fontsize{9pt}{8pt}\selectfont\ttfamily\bfseries,
    columns = fullflexible,
    aboveskip=1pt,
    belowskip=1pt,
    breaklines = true,
    captionpos = b,
    commentstyle = \color{codeblue},
    keywordstyle = \color{codekw},
}

\definecolor{green}{HTML}{009000}
\definecolor{red}{HTML}{ea4335}

\title{RynnVLA-002: A Unified Vision-Language-Action and World Model}

\author[1, 2, 3, *]{Jun Cen}
\author[1, 2, 3, *]{Siteng Huang}
\author[1, 3, *]{Yuqian Yuan}
\author[1, 2, *]{Kehan Li}
\author[1, 2, 3]{Hangjie Yuan}
\author[1]{Chaohui Yu}
\author[1]{Bohan Hou}
\author[1]{Yuming Jiang}
\author[1]{Jiayan Guo}
\author[1, 2]{Xin Li}
\author[1, 2]{Hao Luo}
\author[1]{Fan Wang}
\author[1, 2]{Deli Zhao}
\author[3]{Hao Chen}

\affiliation[1]{DAMO Academy, Alibaba Group}
\affiliation[2]{Hupan Lab}
\affiliation[3]{Zhejiang University}

\abstract{
We introduce RynnVLA-002, a unified Vision-Language-Action (VLA) and world model. The world model leverages action and visual inputs to predict future image states, learning the underlying physics of the environment to refine action generation. Conversely, the VLA model produces subsequent actions from image observations, enhancing visual understanding and supporting the world model’s image generation. The unified framework of RynnVLA-002 enables joint learning of environmental dynamics and action planning. Our experiments show that RynnVLA-002 surpasses individual VLA and world models, demonstrating their mutual enhancement. We evaluate RynnVLA-002 in both simulation and real-world robot tasks. RynnVLA-002 achieves 97.4\% success rate on the LIBERO simulation benchmark without pretraining, while in real-world LeRobot experiments, its integrated world model boosts the overall success rate by 50\%.
}

\date{\today} 
\code{\url{https://github.com/alibaba-damo-academy/RynnVLA-002}}
\correspondence{cenjun.cen@alibaba-inc.com}

\begin{document}
\thispagestyle{firstheader}
\maketitle
\pagestyle{empty}

\section{Introduction}

The Vision-Language-Action (VLA) model has emerged as a promising paradigm for grounding language-conditioned decision making in visual environments, enabling robots to map instructions and observations to actions~\citep{zitkovich2023rt, kim2024openvla}. These models are constructed by augmenting large-scale pre-trained Multimodal Large Language Models (MLLMs) \citep{liu2023visual, li2024llava, zhang2025videollama, bai2025qwen2} with either an action head or additional action expert module to generate actions. MLLMs contribute robust capabilities in perception and decision making, enabling VLA models to exhibit enhanced generalization across a wide range of robotic tasks \citep{black2024pi_0, intelligence2504pi0}.

However, standard VLA architectures face three fundamental drawbacks. First, they cannot fully understand actions because actions reside only on the output side, preventing the model from forming an explicit internal representation of action dynamics. Second, they lack imagination: they do not predict how the world might evolve given candidate actions, hindering foresight and counterfactual reasoning. Third, they have no explicit understanding of physics. Without capturing physical dynamics, the model cannot internalize object interactions, contact, or stability. World models directly address these limitations by learning to forecast future observations conditioned on current images and actions, providing agents with action-aware internal states, imagination, and a physics-informed representation of environment dynamics.~\citep{ha2018world, wu2025ivideogpt}. Despite this advantage, world models are constrained by their inability to directly generate action outputs, resulting in a functional gap that limits their application in scenarios requiring explicit action planning.

\begin{figure*}[t]
  \centering
  \includegraphics[width=0.98\linewidth]{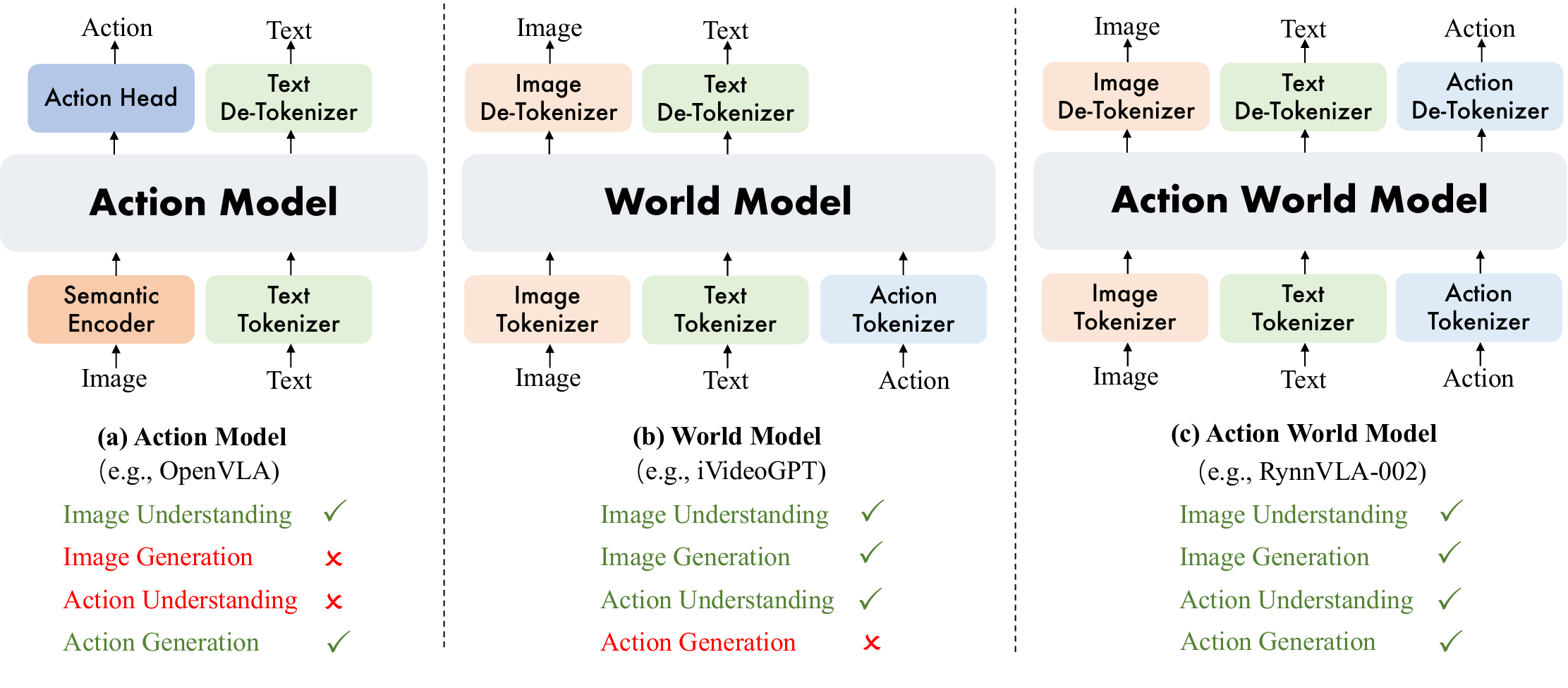}
   \caption{(a) VLA model generates actions based on image understanding; (b) World model generates the image based on image and action understanding; (c) Action World Model unifies both image and action understanding and generation.}
   \label{fig:action_world_model}
\end{figure*}

To address the constraints inherent in both VLA models and world models, we introduce RynnVLA-002, an autoregressive action world model for unified action and image understanding and generation. As depicted in Fig.~\ref{fig:action_world_model}, RynnVLA-002 employs separate tokenizers to encode images, text, states, and actions. The tokens from different modalities are set to share the same vocabulary so that understanding and generation across these modalities can be unified within a single LLM architecture. The world model component captures the underlying physical dynamics of the environment by generating visual representations based on input actions. This process of action interpretation and environmental physics learning is essential for enabling effective decision making within the VLA model. At the same time, the VLA model embedded within RynnVLA-002 refines the understanding of visual data, thereby improving the precision of image generation performed by the world model. Rather than using a world model only as an external reward model, simulator, or pretraining signal, RynnVLA-002 makes action prediction and action-conditioned visual prediction share the same modeling space, enabling the two capabilities to improve each other inside one framework. This bidirectional enhancement creates a more robust and comprehensive model capable of understanding and generating both actions and images.

In this work, we explore different action generation mechanisms. Our initial approach (i.e., WorldVLA~\citep{cen2025worldvla}) involved discretizing actions and unifying them with image and text tokens within a single vocabulary. However, we find that this method struggles with generating coherent action chunks. The primary reason for this is that pretrained multimodal language models have predominantly been exposed to images and text rather than actions, resulting in limited action generalization capabilities. Furthermore, in autoregressive models where subsequent actions are conditioned on preceding ones, error propagation becomes a critical issue, as earlier incorrect predictions influence subsequent actions over time. To alleviate this issue, we proposed an action attention masking strategy that selectively masks prior actions during the generation of current actions. This approach effectively mitigates error accumulation and yields substantial improvements in the task of action chunk generation in simulation.

However, in real-world robot experiments, this discrete design exhibits limited generalization capability and slow inference. We attribute the poor generalization to the high-volume data requirement of discrete autoregressive models~\citep{kaplan2020scaling}, which is often unavailable in robotics. The slow inference, meanwhile, stems from the sequential nature of the autoregressive generation process. To address these issues, we evolve our architecture into a hybrid model that retains the original discrete joint modeling while incorporating a continuous Action Transformer head~\citep{zhao2023learning}. This new head is significantly smaller than the base LLM, which alleviates overfitting and improves generalization. Furthermore, the Action Transformer's parallel decoding and bidirectional attention mechanism reduce the number of decoding steps, accelerating inference, and generating smoother trajectories.

In summary, our contributions are as follows:
\begin{itemize}
    \item We propose RynnVLA-002, an action world model that unifies VLA and World Model in a single framework.
    \item We introduce an action attention masking strategy for the discrete action chunk generation, addressing the challenge of action error accumulation when autoregressively generating action sequences. An additional continuous Action Transformer head is added for stronger generalization ability and smoother trajectory.
    \item Our experiments demonstrate that RynnVLA-002 outperforms the standalone VLA and world models, highlighting the mutual enhancement between the world model and VLA model. Additionally, RynnVLA-002 achieves 97.4\% success rate on the LIBERO simulation benchmark without robot-data pretraining, while in real-world LeRobot experiments, its integrated world model boosts the overall success rate by 50\%.
\end{itemize}

\section{Related Work}

\begin{figure*}[t]
  \centering
  \includegraphics[width=\linewidth]{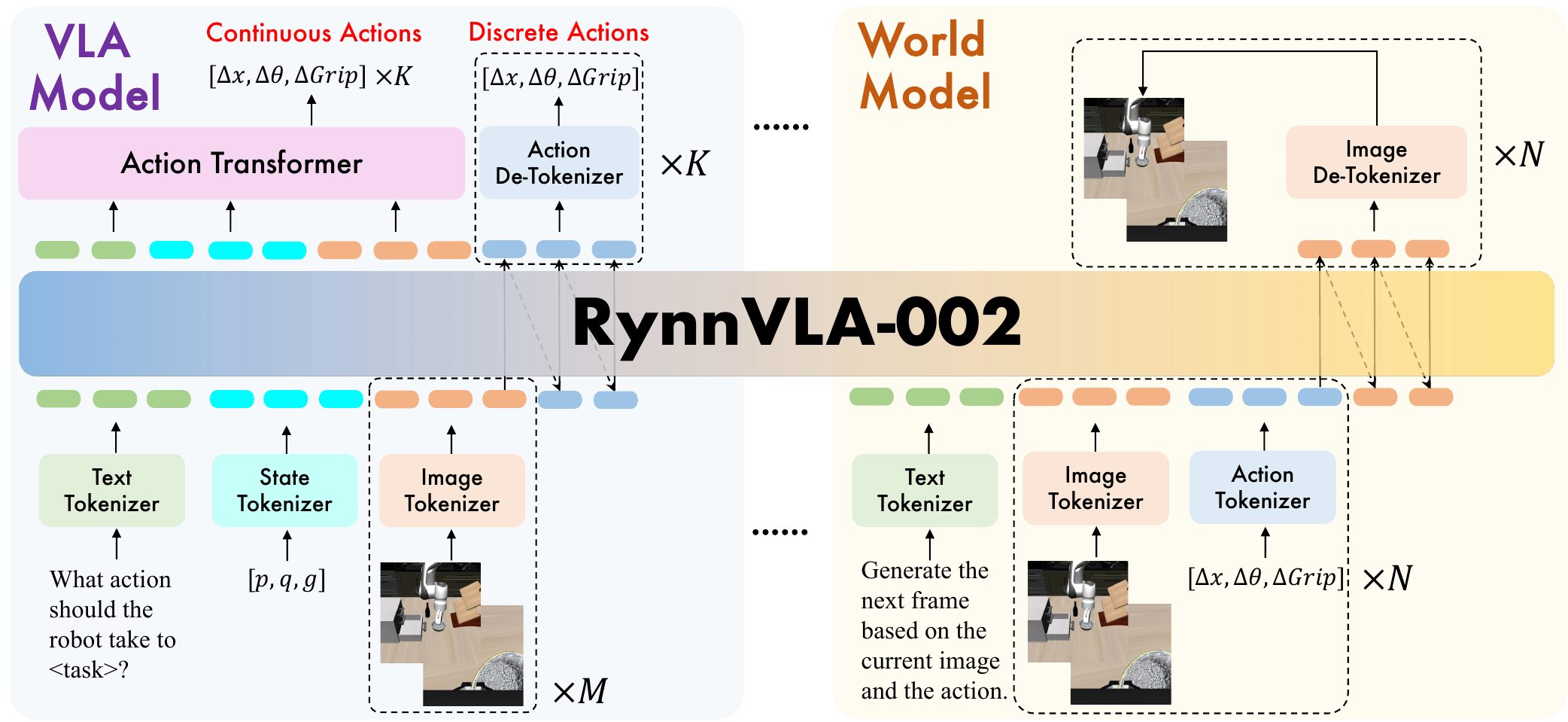}
   \caption{Overview of RynnVLA-002. RynnVLA-002 involves VLA model data and world model data during the training process.}
   \label{fig:outline}
\end{figure*}

\subsection{Vision-Language-Action Models}

\noindent\textbf{VLM-based VLA.} 
Vision-Language Model (VLM)-based VLA models~\citep{brohan2022rt,cheang2024gr,wen2025tinyvla,li2023vision,huang2023embodied,belkhale2024minivla,zhao2025vlas,wang2025vq,wang2025vlaadapter} map visual–language inputs to actions. RT-2~\citep{zitkovich2023rt} first co-trained VLMs on robotic trajectories and web-scale vision-language data, producing actions as discrete tokens. Subsequent works~\citep{wu2023unleashing,kim2024openvla,li2024llara,zhen20243d,pertsch2025fast} extend this architecture to enhance generalization and representation efficiency. To address precision loss from discrete tokens, LCB~\citep{shentu2024llms} introduced a dual-system with a continuous policy head, inspiring variants with different policy head models like diffusion transformers~\citep{peebles2023scalable}, and incorporating diverse
training strategies across multiple embodiments~\citep{zhang2024hirt,wen2024diffusion,wen2025dexvla,zhou2025chatvla,li2024cogact}. Recent frameworks such as $\pi_0$~\citep{black2024pi_0} leverage conditional flow matching~\citep{lipman2022flow}, the open-source GR00T~\citep{bjorck2025gr00t} scales it to complex humanoid control, and $\pi_{0.5}$~\citep{intelligence2504pi0} further improves generalization by leveraging large-scale multimodal web and cross-embodiment data, enabling direct zero-shot deployment across robot platforms. 

\noindent\textbf{Visual Generation-based VLA.} 
Beyond static perception, visual-generation approaches model dynamics by predicting future visual states. UniPi~\citep{du2023learning}, DREAMGEN~\citep{jang2025dreamgen} and GeVRM~\citep{zhang2025gevrm} generate future visual observations to guide action generation.
Joint frameworks~\citep{guo2024prediction,zheng2025flare,li2025unified} co-generate future frames and corresponding actions, enhancing temporal consistency and policy learning. Others exploit future video prediction as a powerful pretraining objective, including GR-2~\citep{cheang2024gr}, VPP~\citep{hu2024video} and RynnVLA-001~\citep{jiang2025rynnvla}. Collectively, these approaches highlight the potential of predictive visual modeling to bridge perception and action, though challenges remain in visual fidelity, domain transfer, and computational efficiency. Our RynnVLA-002 is built on Chameleon~\citep{team2024chameleon}, a unified model for image understanding and generation, thus combining the benefits of both VLM and visual-generation-based approaches.

\noindent\textbf{World-model-assisted VLA.}
Several recent methods also investigate how world models can help VLA policies. 3D-VLA~\citep{zhen20243d} introduces a 3D generative world model to provide additional predictive supervision for VLA learning. IRL-VLA~\citep{jiang2025irlvla}, World-Env~\citep{xiao2025worldenv}, and World-VLA-Loop~\citep{liu2026worldvlaloop} use world models as reward models, virtual simulators, or closed-loop refinement environments for policy post-training. These systems usually keep the policy model and the world model as separate components, using predicted rollouts or rewards to further optimize a VLA. In contrast, RynnVLA-002 internalizes the VLA objective and the action-conditioned world-modeling objective in one Chameleon-style autoregressive backbone with a shared token space, allowing the same model to be queried for action prediction or future visual prediction.

\subsection{World Models}
World models endow embodied AI with internal representations~\citep{chen2022transdreamer,robine2023transformer,wang2024worlddreamer} and predictive dynamics of the external world~\citep{hafner2019dream,hafner2020mastering,okada2022dreamingv2}, enabling physics-consistent interaction in dynamic environments~\citep{xiang2023language,mazzaglia2024genrl,ha2018world}. Recent advances have realized world models with transformer-based architectures~\citep{wu2025paragraph,robine2023transformer,micheli2022transformers}. Notably, Google’s Genie framework~\citep{bruce2024genie} constructs synthetic interactive environments through large-scale self-supervised video pretraining. Nowadays, such world models are widely utilized to generate varied training data~\citep{agarwal2025cosmos}, support model-based reinforcement learning algorithms~\citep{wu2025ivideogpt}, and aid in selecting the most suitable policies from a pool of generated options~\citep{li2025unified,bar2024navigationworldmodels}. In this work, we show that world model and VLA could enhance each other.

\section{Methods}
\subsection{Overview}
The overall architecture of RynnVLA-002 is shown in Fig.~\ref{fig:outline}. As can be seen, our RynnVLA-002 unifies two foundational models in embodied AI. The first is the VLA model, where a policy $\pi$ generates an action $a_t$ based on a language goal $l$, a proprioceptive state $s_{t-1}$, and an observation history $o_{t-h:t}$:
\begin{equation}
    a_t \sim \pi(a_t \mid l, s_{t-1}, o_{t-h:t}).
\end{equation}
The second is the world model, where the model $f$ predicts the next observation $o_t$ from past observations and actions:
\begin{equation}
    \hat{o}_t \sim f(o_t \mid o_{t-h:t-1}, a_{t-h:t-1}).
\end{equation}
We mix the VLA model data and the world model data to train the RynnVLA-002, an integrated model $M_\psi$ that consolidates the capabilities of action prediction and world modeling. The dual nature of our model is captured by its ability to be queried either as a VLA or as a world model, leveraging a shared group of parameters $\psi$.

The term \textit{action world model} refers to this functional unification rather than to reusing the exact modules of a conventional VLA model. A standalone VLA typically relies on a semantic visual encoder and an action head to map observations to actions, while our model uses Chameleon's image tokenizer and decoder so that image observations can be both understood and generated. RynnVLA-002 therefore belongs to the action world model family: it shares a backbone for action prediction and action-conditioned visual prediction, and augments the shared representation with a lightweight continuous Action Transformer for real-world control.


\subsection{Data Tokenization}
 \textbf{Tokenizers.} We initialize the model from Chameleon~\citep{team2024chameleon} since it is a unified model for image understanding and generation. Four tokenizers are involved, including an image tokenizer, a text tokenizer, a state tokenizer, and an action tokenizer.
The image tokenizer is a VQ-GAN model~\citep{esser2021taming} with additional perceptual losses to specific image regions, \textit{e.g.}, faces and salient objects~\citep{gafni2022make}. The compression ratio of the image tokenizer is 16 and the codebook size is 8192. The image tokenizer generates 256 tokens for $256 \times 256$ images and 1024 tokens for $512 \times 512$ images.
The text tokenizer is a trained BPE tokenizer~\citep{sennrich2015neural}. The image and text tokenizers are inherited from Chameleon.
The state and action tokenizer discretizes each dimension of continuous robot proprioceptive states and actions into one of 256 bins, with bin widths determined by the range of the training data~\citep{kim2024openvla, zitkovich2023rt}. All image, text, action, and state tokens are in a single token vocabulary with the size of 65536. The continuous actions generated by the Action Transformer are raw actions without tokenization.

\noindent \textbf{VLA Model Data.} The overall token sequence of VLA model data is:
\begin{center}
$
    \texttt{\{text\}}
    \,
    \texttt{\{state\}}
    \,
    \underbrace{\texttt{\{image-front-wrist\}}}_{\times M}
    \,
\overbrace{
    \underbrace{\texttt{\{action\}}}_{\times K}
}^{\mathcal{L}_{dis\_action}}
$
\end{center}
The VLA model generates $K$ actions based on the language instruction, proprioceptive state, and $M$ historical image observations. The text inputs are \textit{``What action should the robot take to + \textless task\textgreater + ?''}. $\mathcal{L}_{dis\_action}$ refers to the cross-entropy loss of discrete action tokens.

\noindent \textbf{World Model Data.} World model generates the next image frame given the current image observation and action. The overall token sequence is:
\begin{center}
\texttt{\{text\}}
$\underbrace{ 
  \texttt{\{images-front-wrist\}\{action\}}
    \overbrace{\texttt{\{images-front-wrist\}} 
}^{\mathcal{L}_{img}}}_{\times N}$ 
\end{center}
All of the training instances for world model share the same text prefix \textit{``Generate the next frame based on the current image and the action.''} and there are no other task instructions since the action could totally determine the next state of the world. The overall generation could repeat $N$ times in an autoregressive manner. $\mathcal{L}_{img}$ refers to the cross-entropy loss of discrete image tokens.
\begin{figure}[t]
  \centering
\includegraphics[width=\linewidth]{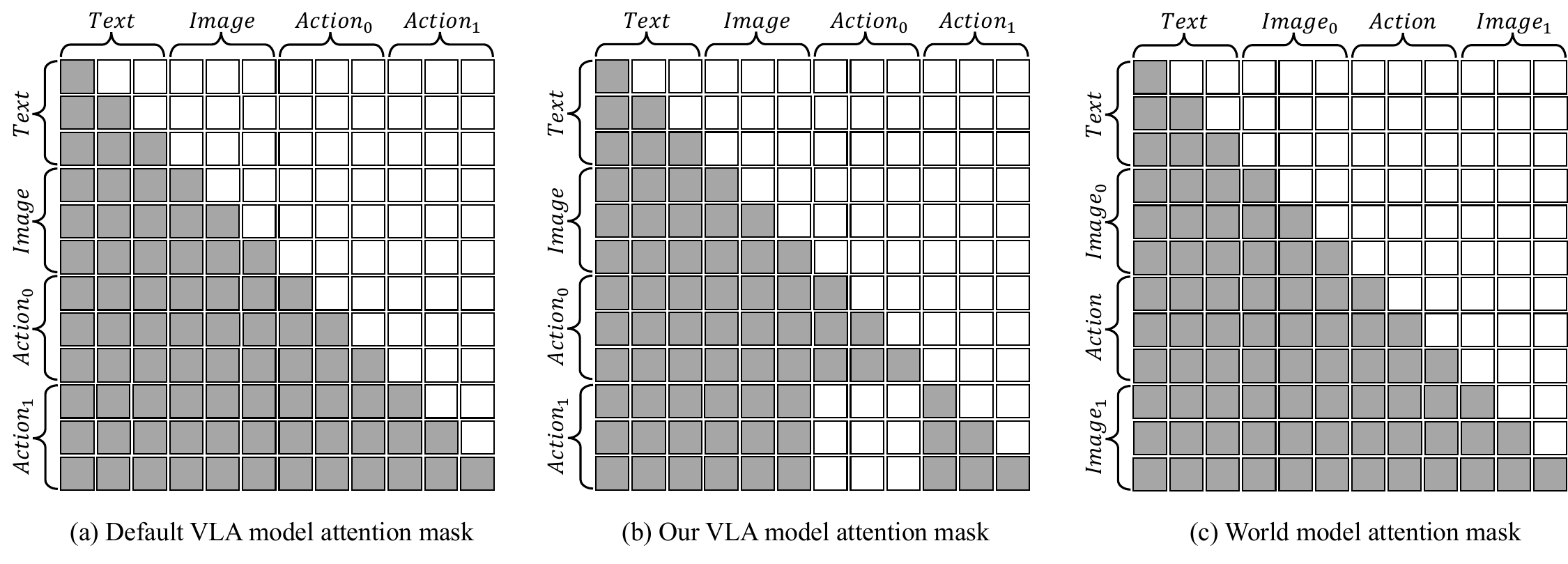}
   \caption{Attention mask of (a) default VLA model, (b) our proposed VLA model, and (c) world model. A gray block means the token in the corresponding row can attend to the token in the corresponding column.}
   \label{fig:attention_mask}
\end{figure}

\noindent \textbf{Training Objective.} We mix the VLA model data and world model data to train our RynnVLA-002. Unless otherwise specified, the two task types are sampled with a 1:1 ratio, and the model is trained in a single finetuning stage without additional large-scale robot manipulation pretraining. The overall loss function is $\mathcal{L}_{dis}=\mathcal{L}_{dis\_action}+\mathcal{L}_{img}$. In this way, RynnVLA-002 could behave as the VLA model or world model depending on the user queries.

\noindent \textbf{Inference Modes.} Although RynnVLA-002 supports both action prediction and future-image prediction, it performs only one queried task during a single inference call. When used as a VLA policy, the model consumes the language instruction, proprioceptive state, and visual history to generate actions through either discrete action tokens or the continuous Action Transformer; it does not roll out future images for control. When used as a world model, the model predicts future visual observations conditioned on images and actions. This decoupled querying avoids introducing autoregressive image-prediction compounding errors into the real-time action generation loop.

\subsection{Action Chunk Generation}
\textbf{Attention Mask for Discrete Action Chunk.} Generating multiple actions for execution is critical for efficiency and a higher success rate~\citep{kim2025fine}. However, we find that naively generating consecutive actions in the autoregressive model degrades the performance. Although the foundational MLLM demonstrates robust generalization capabilities across the image and text domains, its capacity to generalize effectively in the action domain is comparatively limited. Consequently, errors originating from earlier actions propagate to subsequent actions under the default causal attention mask, resulting in performance degradation. To address this limitation, we introduce an alternative attention mask tailored for action generation, depicted in Fig.~\ref{fig:attention_mask} (b). This modified mask ensures that current actions rely solely on textual and visual input, while prohibiting access to prior actions. Such design enables the autoregressive framework to generate multiple actions in isolation, mitigating the error accumulation problem. The world model part adheres to the conventional attention mask, as shown in Fig.~\ref{fig:attention_mask} (c).

\begin{table*}[t!]
\centering
\setlength{\tabcolsep}{1.74mm} 
\caption{Evaluation results on LIBERO benchmark. Pretraining means the model is pretrained on the large-scale robot manipulation data.}
\begin{tabular}{lccccccc}
\toprule
\textbf{Methods} & Pretraining &Action Type & \begin{tabular}[c]{@{}c@{}}Spatial \end{tabular} & \begin{tabular}[c]{@{}c@{}}Object \end{tabular} & \begin{tabular}[c]{@{}c@{}}Goal \end{tabular} & \begin{tabular}[c]{@{}c@{}}Long \end{tabular} & \begin{tabular}[c]{@{}c@{}}Average \end{tabular} \\ \midrule
LAPA~\citep{ye2024latent} & \ding{55} &Discrete & 73.8 & 74.6 & 58.8 & 55.4 & 65.7 \\
TraceVLA~\citep{zheng2024tracevla} & \ding{51} &Discrete & 84.6 & 85.2 & 75.1 & 54.1 & 74.8 \\
OpenVLA~\citep{kim2024openvla} & \ding{51} &Discrete & 84.7 & 88.4 & 79.2 & 53.7 & 76.5 \\
SpatialVLA~\citep{qu2025spatialvla} & \ding{51} &Discrete & 88.2 & 89.9 & 78.6 & 55.5 & 78.1 \\
NORA~\citep{hung2025nora} & \ding{55} &Discrete & 85.6 & 89.4 & 80.0 & 63.0 & 79.5 \\
CoT-VLA~\citep{zhao2025cot} & \ding{51} &Discrete & 87.5 & 91.6 & 87.6 & 69.0 & 83.9 \\
$\pi_0$-FAST~\citep{black2024pi_0} & \ding{51} &Discrete & 96.4 & 96.8 & 88.6 & 60.2 & 85.5 \\
MolmoAct~\citep{lee2025molmoact} & \ding{51} &Discrete & 87.0 & 95.4 & 87.6 & 77.2 & 86.6 \\
FlowVLA~\citep{zhong2025flowvla} & \ding{55} &Discrete & 93.2 & 95.0 & 91.6 & 72.6 & 88.1 \\
UniVLA~\citep{bu2025learning}& \ding{51} &Discrete & 96.5 & 96.8 & 95.6 & 92.0 & 95.2 \\ \midrule
Diffusion Policy~\citep{chi2023diffusion} & \ding{55} &Continuous & 78.3 & 92.5 & 68.3 & 50.5 & 72.4 \\
Octo~\citep{team2024octo} & \ding{51} &Continuous & 78.9 & 85.7 & 84.6 & 51.1 & 75.1 \\
MDT~\citep{reuss2024multimodal} & \ding{55} &Continuous & 78.5 & 87.5 & 73.5 & 64.8 & 76.1 \\
DiT Policy~\citep{hou2024diffusion} & \ding{51} &Continuous & 84.2 & 96.3 & 85.4 & 63.8 & 82.4 \\
MaIL~\citep{jia2024mail} & \ding{55} &Continuous & 74.3 & 90.1 & 81.8 & 78.6 & 83.5 \\
ThinkAct~\citep{huang2025thinkact} & \ding{51} &Continuous & 88.3 & 91.4 & 87.1 & 70.9 & 84.4 \\
$\pi_0$~\citep{black2024pi_0}& \ding{51} &Continuous & 90.0 & 86.0 & 95.0 & 73.0 & 86.0 \\
SmolVLA~\citep{shukor2025smolvla}& \ding{55} &Continuous & 93.0 & 94.0 & 91.0 & 77.0 & 88.8 \\
$\pi_{0.5}$~\citep{intelligence2504pi0} & \ding{51} &Continuous & 98.8 & 98.2 & \underline{98.0} & 92.4 & 96.9 \\
OpenVLA-OFT~\citep{kim2025fine} & \ding{51} &Continuous & 97.6 & 98.4 & 97.9 & 94.5 & 97.1 \\
X-VLA~\citep{zheng2025xvla} & \ding{51} &Continuous & 98.2 & 98.6 & 97.8 & \textbf{97.6} & \underline{98.1} \\
EO-1~\citep{qu2025eo1} & \ding{51} &Continuous & \textbf{99.7} & \textbf{99.8} & \textbf{99.2} & \underline{94.8} & \textbf{98.2} \\
Seer~\citep{tian2024predictive} & \ding{51} &Continuous & -- & -- & -- & 87.7 & -- \\ 
UVA~\citep{li2025unified} & \ding{55} &Continuous & -- & -- & -- & 93.0 & -- \\
\midrule
RynnVLA-002-Discrete & \ding{55} &Discrete &94.2 & 96.8 &94.6 &87.6 &93.3 \\
RynnVLA-002-Continuous& \ding{55} &Continuous &\underline{99.0} & \textbf{99.8} & 96.4 &94.4 &97.4 \\
\bottomrule
\end{tabular}
\label{tab:bench_combined}
\end{table*}

\noindent \textbf{Action Transformer for Continuous Action Chunk.} Although our discrete action chunking model performs well in simulation, it rarely succeeds in real-world robot experiments. This discrepancy arises because real-world applications demand significantly higher generalization to cope with dynamic variables like lighting and object positioning—factors not fully captured in simulation. The failure of our discrete model is rooted in two key issues. First, its large autoregressive architecture is prone to severe overfitting when trained on limited real-world dataset, leading to poor generalization. Second, our designed attention mask makes the autoregressive model generate each action in isolation within the same chunk, which cannot ensure trajectory continuity, resulting in severe shaking and non-smooth movements that drastically reduce the success rate.

To overcome these challenges, we propose to augment our architecture with a dedicated Action Transformer to generate continuous action sequences~\citep{zhao2023learning}. This module processes the full context—including language, image, and state tokens—and utilizes learnable action queries to output an entire action chunk in parallel. This design provides two distinct advantages. First, the Action Transformer's more compact architecture is less prone to overfitting on limited data, thereby improving generalization and producing fluid, stable actions. Second, by parallelly generating all actions in a single forward pass, it substantially accelerates the inference speed compared to the autoregressive baselines that generate actions sequentially. We use L1 regression loss $\mathcal{L}_{conti\_action}$ to supervise the Action Transformer. The overall loss function is $\mathcal{L}=\mathcal{L}_{dis}+\alpha\mathcal{L}_{conti}=\mathcal{L}_{dis\_action}+\mathcal{L}_{img}+\alpha\mathcal{L}_{conti\_action}$.

\section{Experiments}

\subsection{Simulation Results}
\textbf{Benchmark.}
We evaluate our method on the LIBERO benchmark~\citep{liu2023libero}. This benchmark is composed of four distinct suites designed to test a range of robotic capabilities: (1) LIBERO-Spatial focuses on spatial relationships by tasking the robot with placing a bowl in various locations; (2) LIBERO-Object emphasizes object recognition and manipulation with unique objects; (3) LIBERO-Goal tests procedural learning by varying task goals while using a fixed set of objects; (4) LIBERO-Long contains 10 complex long-horizon manipulation tasks.

\noindent \textbf{Dataset and Preprocessing.}
We first curate the dataset by removing unsuccessful trajectories and filtering out ``no-operation'' actions, a procedure also adopted by OpenVLA~\citep{kim2024openvla}. For world model evaluation, which relies on ground-truth video-action pairs, we partition the cleaned data into a 90\% training set and a 10\% validation set.

\noindent \textbf{Hyperparameter Settings.}
The VLA model takes $M=2$ historical image frames as input. We set the action chunk size $K=10$ for the longer LIBERO-Long and LIBERO-Spatial tasks and $K=5$ for the shorter LIBERO-Object and LIBERO-Goal tasks. For the world model, we use a single prediction round ($N=1$) to maintain computational efficiency. The loss weighting parameter $\alpha=10$.
\begin{wrapfigure}{r}{0.5\textwidth}
  \centering
  \includegraphics[width=\linewidth]{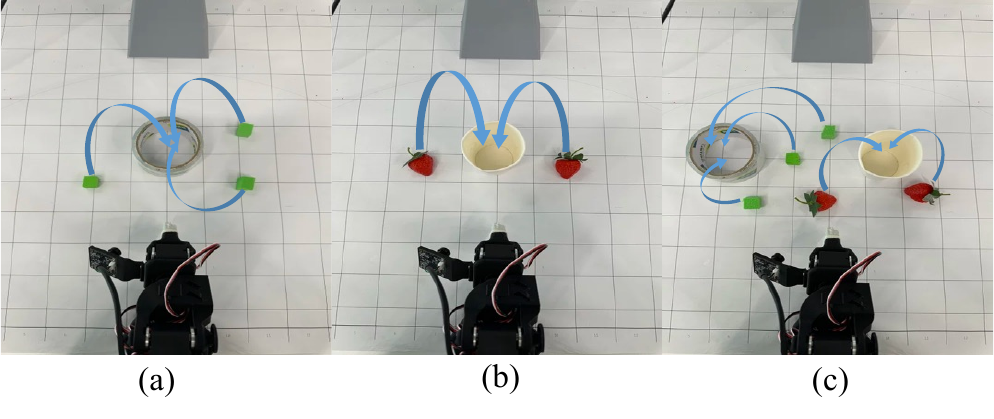} 
  \vspace{-0.6cm}
  \captionof{figure}{Real-world robot settings. (a) Place the block inside the circle. (b) Place the strawberries into the cup. (c) Task with distractors.}
  \label{fig:real_robot_setting}
  \vspace{-0.5cm} 
\end{wrapfigure}
\noindent \textbf{Metrics.}
Our evaluation is twofold. To assess the VLA model, we measure its success rate across 50 deployment rollouts per task, each initialized in a different state. To assess the world model, we measure its video prediction accuracy on the held-out validation set using four standard metrics: Fréchet Video Distance (FVD), Peak Signal-to-Noise Ratio (PSNR), Structural Similarity Index (SSIM), and Learned Perceptual Image Patch Similarity (LPIPS).

\noindent \textbf{Benchmark Results.} We evaluate the performance of discrete actions and continuous actions separately. As shown in Tab.~\ref{tab:bench_combined}, our RynnVLA-002 achieves high success rates of 93.3\% with discrete actions and 97.4\% with continuous actions, demonstrating the effectiveness of our core design principles: jointly learning the VLA modeling and world modeling, an attention mask mechanism for discrete action generation, and the added continuous Action Transformer. We additionally compare with recent strong policies including $\pi_{0.5}$, X-VLA, and EO-1. These methods benefit from large-scale robot or cross-embodiment pretraining, whereas RynnVLA-002 is trained without such robot pretraining. Under this setting, RynnVLA-002 remains competitive with these pretrained baselines and outperforms most non-pretrained methods, indicating that the unified action-world objective provides a strong learning signal even when large external robot corpora are unavailable.

\begin{table*}[t]
\centering
\setlength{\tabcolsep}{3 mm}
\caption{Evaluation results on real-world SO100 robots. Success rate is reported.}
    \begin{tabular}{lcccc}
    \toprule
    \textit{Place the block inside the circle.} & Pretraining & \multicolumn{1}{l}{Single-Target} & \multicolumn{1}{l}{Multi-Target} & \multicolumn{1}{l}{w/ Distractors} \\
    \midrule
    GR00T N1.5~\citep{bjorck2025gr00t} & \ding{51} & 90.0 & 60.0 & 50.0 \\
    $\pi_0$~\citep{black2024pi_0} & \ding{51} & 100.0 & 70.0 & 50.0 \\
    $\pi_{0.5}$~\citep{intelligence2504pi0} & \ding{51} & 90.0 & 60.0 & 80.0 \\
    RynnVLA-002 & \ding{55} & 90.0 & 90.0 &80.0 \\
    \midrule
    \textit{Place the strawberries into the cup.} & Pretraining & \multicolumn{1}{l}{Single-Target} & \multicolumn{1}{l}{Multi-Target} & \multicolumn{1}{l}{w/ Distractors} \\
    \midrule
    GR00T N1.5~\citep{bjorck2025gr00t} & \ding{51} & 50.0 & 50.0 & 70.0 \\
    $\pi_0$~\citep{black2024pi_0} & \ding{51} & 80.0 & 70.0 & 40.0 \\
    $\pi_{0.5}$~\citep{intelligence2504pi0} & \ding{51} & 70.0 & 70.0 & 50.0 \\
    RynnVLA-002 & \ding{55} & 80.0 & 80.0 &50.0 \\
\bottomrule
\end{tabular}
\label{tab:real_robot}
\vspace{-0.3cm}
\end{table*}


\subsection{Real-World Robot Results}

\noindent \textbf{Datasets.} We curate a new real-world manipulation dataset collected with a LeRobot SO100 robotic arm~\citep{cadene2024lerobot}. All trajectories are expert demonstrations obtained via human teleoperation. We define two pick and place tasks for evaluation.
(1) Place the block inside the circle: Emphasizing basic object detection and grasp execution (248 demonstrations); (2) Place strawberries in the cup: Requiring fine-grained localization and grasp-point prediction (249 demonstrations).

\noindent \textbf{Baselines.} We compare with three strong open-source baselines: GR00T N1.5~\citep{bjorck2025gr00t}, $\pi_0$~\citep{black2024pi_0}, and $\pi_{0.5}$~\citep{intelligence2504pi0}. For these methods, we initialize from the official pretrained checkpoints and finetune them on the same SO100 dataset used for our model. We adopt the same recipe from the official codebases of these baselines to do finetuning.

\noindent \textbf{Evaluation.} As shown in Fig.~\ref{fig:real_robot_setting}, our evaluation spans three scenarios: (1) {Single-target manipulation}, with exactly one target object on the desktop; (2) {Multi-target manipulation}, with multiple target objects present; and (3) {Instruction-following with distractors}, where both targets and distractors appear. A trial is deemed successful if the robot places at least one target object into its designated location within a predefined time budget. A trial fails if: (1) the time limit is exceeded; (2) the robot accrues more than five consecutive failed grasp attempts on a target; (3) in the instruction-following with distractors setting, the agent attempts to manipulate any distractor objects. Each task is tested for 10 times and we report the success rate.

\begin{table*}[t]
\caption{Ablation study of VLA model using discrete actions on LIBERO benchmark.
}
\centering
\setlength{\tabcolsep}{2.34 mm}
\begin{tabular}{ccccccccccc}
\toprule
Index&
 \makecell{VLA \\ Discrete} & \makecell{World \\ Model} & \makecell{Action \\ Chunking} & \makecell{Our VLA model\\Attention Mask} & \makecell{Goal} & \makecell{Object} & \makecell{Spatial} & \makecell{Long} & \makecell{Average}\\
\midrule
 1& \ding{51} &\ding{55} &\ding{55} &\ding{55}    & 67.3          & 82.9           & 77.8            & 23.0         &62.8\\
2& \ding{51} &\ding{51} &\ding{55} &\ding{55} & 73.1          & 88.0           & 80.2            & 27.3         &67.2          \\
3& \ding{51} &\ding{55} &\ding{51} &\ding{55} &79.6   &82.9 &36.7 &16.9 &54.0      \\
4& \ding{51} &\ding{55} &\ding{51} &\ding{51} & 84.4 &90.9 &81.8 &49.3  &76.6      \\
5& \ding{51} &\ding{51} &\ding{51} &\ding{51} & 85.1 &90.9 &84.0 &52.4 &78.1     \\
\bottomrule
\label{tab:ab_dis_action}
\end{tabular}
\vspace{-0.6cm}
\end{table*}

\noindent \textbf{Results.} Tab.~\ref{tab:real_robot} shows the results of real-world robot experiments. Our RynnVLA-002 achieves competitive results with GR00T N1.5~\citep{bjorck2025gr00t}, $\pi_0$~\citep{black2024pi_0}, and $\pi_{0.5}$~\citep{intelligence2504pi0} without robot pretraining. $\pi_{0.5}$ improves over $\pi_0$ in distractor-filled scenes, suggesting stronger robustness under visual clutter. RynnVLA-002 still achieves the best overall performance across the six real-world settings. For instance, RynnVLA-002 reaches 90\% success on the multi-target "Place the block" task and 80\% success when distractors are present, improving over the strongest pretrained baseline by 20\% in the multi-target setting and matching its distractor robustness.

\begin{table*}[t!]
\caption{Ablation study of VLA model using continuous actions on LIBERO benchmark.
}
\centering
\setlength{\tabcolsep}{2.3 mm}
\begin{tabular}{ccccccccccc}
\toprule
Index&
 \makecell{VLA \\ Continuous} & \makecell{World \\ Model} & \makecell{Wrist \\ Camera} & \makecell{Proprioceptive\\State} & \makecell{Goal} & \makecell{Object} & \makecell{Spatial} & \makecell{Long} & \makecell{Average}\\
\midrule
1 &\ding{51} & \ding{55} & \ding{55} & \ding{55} & 90.2 &92.4 &88.4 &67.0 &84.5\\
2 &\ding{51} & \ding{55} & \ding{51} & \ding{55} & 91.4 &95.4 &98.2 & 81.4 &91.6\\
3 &\ding{51} & \ding{51} & \ding{51} & \ding{55} & 96.0 &97.4 &99.0 &85.8 &94.6\\
4 &\ding{51} & \ding{51} & \ding{51} & \ding{51} & 96.4 &99.8 &99.0 &94.4 &97.4\\
\bottomrule
\label{tab:ab_conti_action}
\end{tabular}
\vspace{-0.5cm}
\end{table*}

\subsection{Ablation Study}
\noindent \textbf{World Model Benefits the VLA Model.}  
On the LIBERO simulation benchmark, incorporating world model data during training consistently improves performance under the same backbone, data, and training recipe. Specifically, in Tab.~\ref{tab:ab_dis_action}, the success rate for discrete actions increases from 62.8\% (Line 1) to 67.2\% (Line 2), and from 76.6\% (Line 4) to 78.1\% (Line 5). A similar trend is observed for continuous actions in Tab.~\ref{tab:ab_conti_action}, where the success rate increases from 91.6\% (Line 2) to 94.6\% (Line 3). These controlled comparisons isolate the contribution of the world-modeling objective; they are therefore complementary to the benchmark table, where some externally pretrained models use different backbones and much larger robot corpora. In real-world robot experiments, the benefit of world model data is even more pronounced. As shown in Line 4 of Tab.~\ref{tab:ab_real_action}, the model trained without world model data achieves a very low success rate on real-world robot tasks, which is below 30\%. In contrast, augmenting VLA training with world model significantly boosts performance to over 80\%.

\begin{figure*}[t!]
  \centering
  \includegraphics[width=0.98\linewidth]{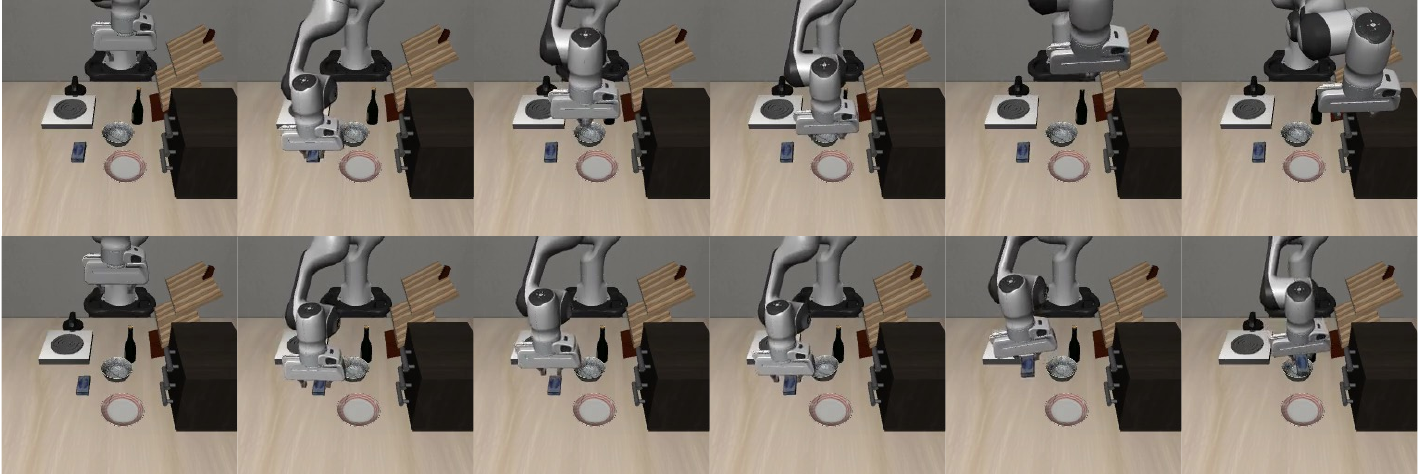}
   \caption{VLA model visualization on LIBERO. Task: put the cream cheese in the bowl. Top: w/o world model. Bottom: w/ world model.}
   \label{fig:vis_vla}
\end{figure*}

Fig.~\ref{fig:vis_vla} shows that the model trained without world model data often lacks target-oriented focus: it may move directly toward the target placement region without successfully grasping the cheese or bottle, and sometimes fails to recover after the first unsuccessful attempt. The model jointly trained with world model keeps retrying to grasp the target object when encountering failures. This behavior is consistent with the world-modeling objective: when predicting the next image, moving objects contribute a larger and harder-to-model portion of the loss than the mostly static background, so the shared representation receives stronger supervision on target objects and their interaction dynamics.

\begin{table}[t!]
\caption{Ablation study of VLA model on real-world robots.}
\label{tab:ab_real_action}
\centering
\setlength{\tabcolsep}{3.5 mm}
\begin{tabular}{c ccc ccc cccc}
\toprule
Index & {\makecell{Action \\ Type}} & {\makecell{World \\ Model}} &\makecell{Wrist \\ Camera} & \makecell{Proprioceptive \\State} & {\makecell{Single \\ Target}} & {\makecell{Multi \\ Target}} &{\makecell{with \\ Distractors}}\\
\midrule
1 & Discrete  & \ding{51} & \ding{51} & \ding{51} &0 &0  &0\\ 
2 & Continuous  & \ding{51} & \ding{51} & \ding{55} &0 &0 &0 \\
3 & Continuous  & \ding{51} & \ding{55} & \ding{51} &0 &0 &0 \\
4 & Continuous  & \ding{55} & \ding{51} & \ding{51} &30.0 &10.0 &0 \\
5 & Continuous  & \ding{51} & \ding{51} & \ding{51} &80.0 &80.0 &50.0\\
\bottomrule
\end{tabular}
\vspace{-0.3cm}
\end{table}

\begin{figure*}[t!]
  \centering
  \includegraphics[width=0.98\linewidth]{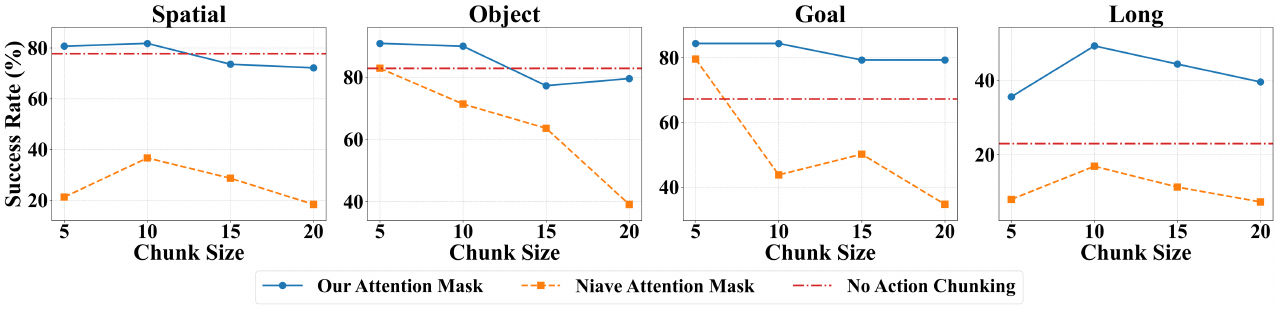} 
  \caption{Ablation study of chunk length with discrete actions.} 
  \label{fig:action_ck}
\end{figure*}

\begin{figure*}[t!]
  \centering  \includegraphics[width=0.98\linewidth]{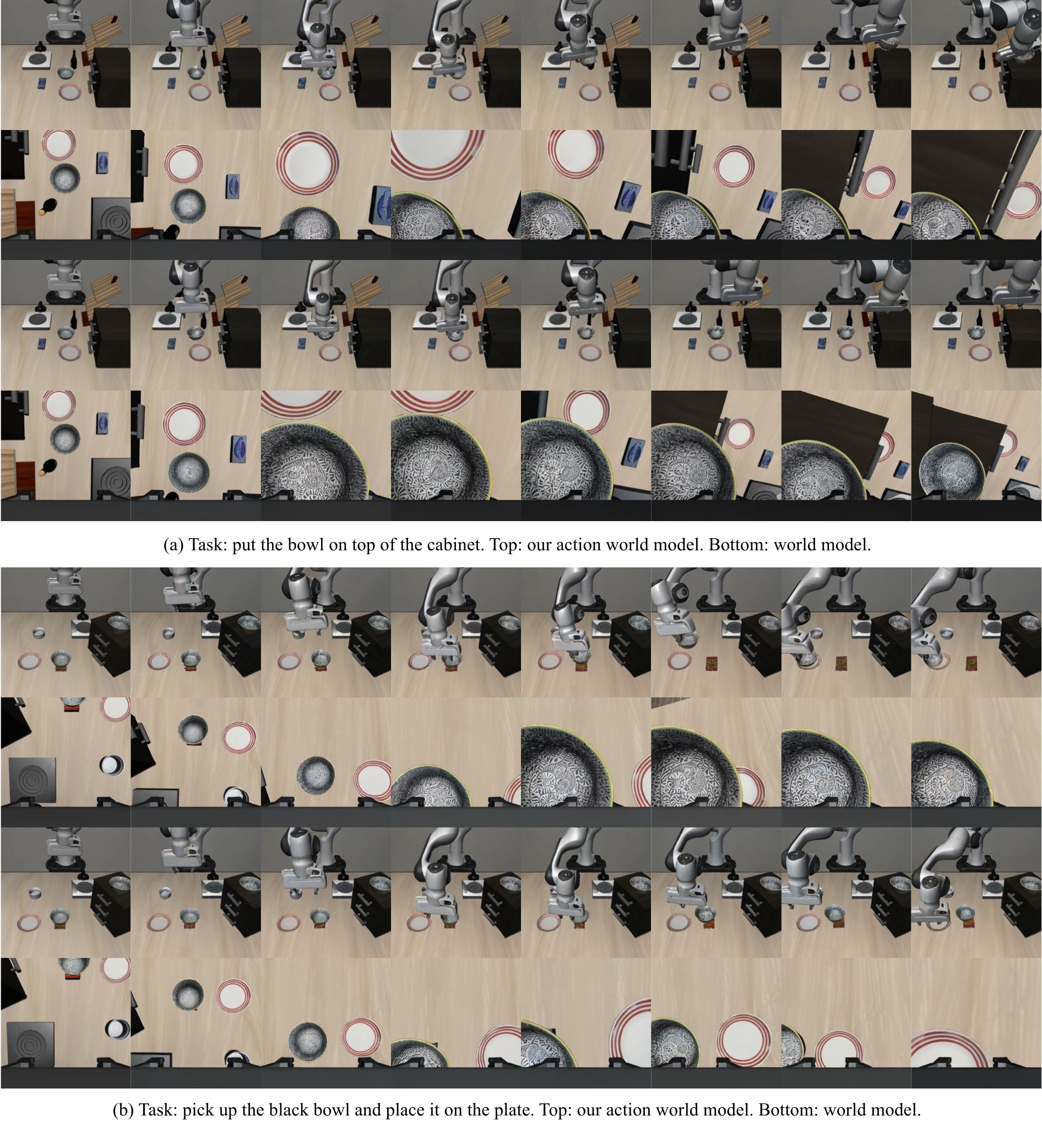}
    \vspace{-0.4cm}
   \caption{World model visualization.}
   \label{fig:vis_wm}
   \vspace{-0.4cm}
\end{figure*}

\noindent \textbf{VLA Model Enhances the World Model.} As shown in Tab.~\ref{tab:ab_world}, the model trained on a mixture of VLA and world model data achieves generation results that are comparable to or better than those of the world model trained solely on world data.
%
Furthermore, Fig.~\ref{fig:vis_wm} compares video generations from our action world model with a baseline world model trained without VLA data. The baseline world model fails to predict a successful grasp of the bowl from the front camera perspective in both two examples. 
\begin{wraptable}[18]{r}{0.52\textwidth}
\caption{Ablation study of world model on LIBERO validation set.}
\label{tab:ab_world}
\centering
\setlength{\tabcolsep}{1mm} 
\begin{tabular}{lcccc}
\toprule
\textit{Goal} & {FVD}$\downarrow$ & {PSNR}$\uparrow$ & {SSIM}$\uparrow$ & {LPIPS}$\downarrow$ \\
\midrule
World Model      & 370.0 & \textbf{22.25}          & 77.84 & 19.70          \\
Action World Model & \textbf{336.8}        & 22.13 & \textbf{78.13}          & \textbf{19.43} \\
\midrule
\textit{Object} & {FVD}$\downarrow$ & {PSNR}$\uparrow$ & {SSIM}$\uparrow$ & {LPIPS}$\downarrow$ \\
\midrule
World Model & 1141.6 &20.31 &59.59 &27.30 \\
Action World Model & \textbf{877.2} & \textbf{22.18} & \textbf{65.03} & \textbf{22.60}\\
\midrule
\textit{Spatial} & {FVD}$\downarrow$ & {PSNR}$\uparrow$ & {SSIM}$\uparrow$ & {LPIPS}$\downarrow$ \\
\midrule
World Model & 405.4 & 22.32 & 79.15 & 20.28\\
Action World Model & \textbf{373.1} & \textbf{23.88} & \textbf{82.41} & \textbf{16.33} \\
\midrule
\textit{Long} & {FVD}$\downarrow$ & {PSNR}$\uparrow$ & {SSIM}$\uparrow$ & {LPIPS}$\downarrow$ \\
\midrule
World Model & 557.73 & 18.24 & 69.16 & 31.60\\
Action World Model &\textbf{427.86} & \textbf{19.36} &\textbf{72.19} &\textbf{27.78}\\
\bottomrule
\end{tabular}
\end{wraptable}
In contrast, our action world model consistently generates correct videos depicting a successful grasp. Notably, the baseline world model also exhibits a significant inconsistency: as seen in Fig.~\ref{fig:vis_wm} (a), the front camera shows a failed grasp while the wrist camera shows a successful one. This highlights a critical disalignment between the model's predictions for different viewpoints. The visualization results validate that the image understanding capabilities inherited from the VLA model strengthen the world model’s generation performance. 

\noindent \textbf{Attention Mask for Discrete Action Chunk Generation.} Simultaneous generation of multiple actions is essential for achieving effective and efficient grasping. However, we observe that a vanilla autoregressive approach—where actions are generated sequentially—can degrade model performance, as evidenced by the results in row 3 of Table~\ref{tab:ab_dis_action} and Fig.~\ref{fig:action_ck}. The grasping success rate gradually decreases with longer action chunks. This degradation arises because later actions become overly dependent on preceding ones since they share the same space, rather than being grounded in visual input which is a distinct modality. The generalization of the action is not that strong as this modality was not involved during pretraining the MLLM. Consequently, errors tend to accumulate as the sequence of generated actions increases. The proposed attention masking mechanism ensures that each action is generated independently and solely determined by the visual input, thereby mitigating the issue of error propagation within the action sequence. 
\begin{wrapfigure}{r}{0.52\textwidth}
  \centering
  \includegraphics[width=\linewidth]{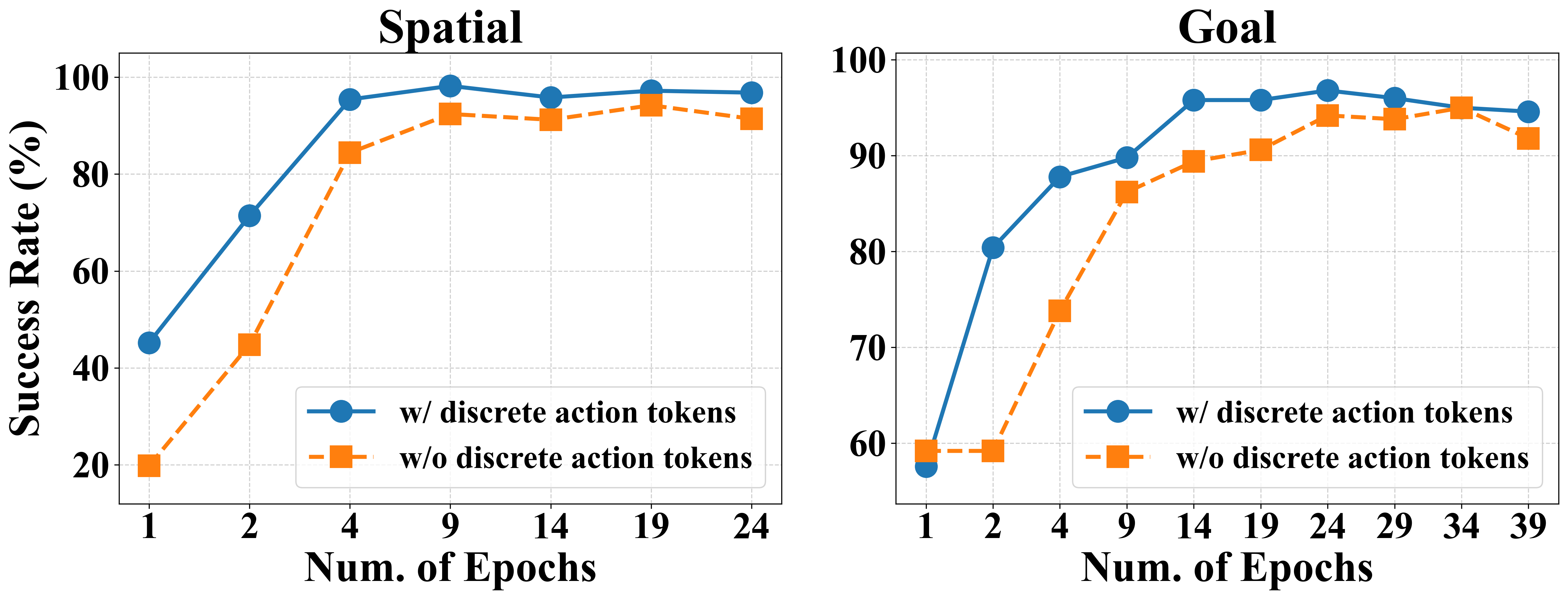} 
  \vspace{-0.4cm}
  \captionof{figure}{Discrete action tokens accelerate the convergence of continuous action generation.}
  \label{fig:real_dis_action}
  \vspace{-0.5cm} 
\end{wrapfigure}

As illustrated in Fig.~\ref{fig:action_ck}, the model incorporating the proposed attention mask demonstrates superior performance compared to the vanilla attention mask, particularly under conditions of longer chunk lengths. This highlights the efficacy of the introduced masking approach. If the length of the action chunk is excessively prolonged, the robot's ability to timely adapt its policy becomes constrained, leading to a decline in overall performance.


\noindent \textbf{Discrete Actions Accelerate the Convergence.} We retain discrete actions during training alongside the continuous Action Transformer, as we find that this hybrid approach not only speeds up the convergence of VLA training 
but also improves the ultimate success rate. 
As shown in Fig.~\ref{fig:real_dis_action}, models trained with discrete action tokens achieve a substantially higher success rate than those trained without them, with the advantage being most pronounced during the initial stages of training.

\noindent \textbf{Ablation Study of Wrist Camera and Proprioceptive State.}  
As shown in Line 1 of Tab.~\ref{tab:ab_conti_action}, our model could achieve reasonable performance without the wrist camera or proprioceptive state on the LIBERO simulation benchmark. Incorporating these two sources of information brings in further performance gains (see Line 2 and Line 4 in Tab.~\ref{tab:ab_conti_action}). In real-world experiments (Line 2 and Line 3 in Tab.~\ref{tab:ab_real_action}), the robot consistently fails when the wrist camera or proprioceptive state is absent. On one hand, the wrist camera provides crucial visual feedback on the relative pose between the gripper and the object, especially when the robot is outside the field of view of the front camera. On the other hand, we find proprioceptive state is essential for accurately timing gripper closure and object lifting during manipulation.

\noindent \textbf{Efficiency Analysis.} As shown in Tab.~\ref{tab:ab_eff}, incorporating additional input images, such as images from the wrist camera or historical frames, improves performance but reduces speed. For discrete actions, action chunking yields both higher inference speed and better performance compared to generating a single action per inference step. Continuous action generation is significantly faster owing to its parallel generation nature, and frequency scale almost linearly with chunk size as generating additional actions incurs negligible extra time.

\begin{table*}[t!]
\caption{Ablation study of VLA model on efficiency and action chunking on LIBERO benchmark. Frequency is measured in Hz.}
\label{tab:ab_eff}
\centering
\setlength{\tabcolsep}{1.55 mm}
\begin{tabular}{cc ccc ccc ccc}
\toprule
 & & & & & \multicolumn{3}{c}{Chunk Size = 5} & \multicolumn{3}{c}{Chunk Size = 10} \\
\cmidrule(lr){6-8} \cmidrule(lr){9-11}
\multirow{-2}{*}{Index} & \multirow{-2}{*}{\makecell{Action \\ Type}} & \multirow{-2}{*}{\makecell{Action \\ Chunking}} & \multirow{-2}{*}{\makecell{Wrist \\ Camera}} & \multirow{-2}{*}{\makecell{History \\ Length}} & Frequency & Goal & Object & Frequency & Spatial & Long \\
\midrule
1 & Discrete  & \ding{55} & \ding{55} & 0 &2.50 &60.0 &71.0 &2.50       &77.0       &10.4      \\
2 & Discrete  & \ding{51} & \ding{55} & 0 &3.69      & 83.2 & 90.0 &3.69       & 83.6  & 46.0 \\ \midrule
3 & Discrete  & \ding{55} & \ding{55} & 1 & 1.88 &74.6 &72.8 &1.88 &78.6 &17.0 \\
4 & Discrete  & \ding{51} & \ding{55} & 1 &3.24      & 92.8 & 91.6 &3.24       & 86.2  & 64.0 \\ \midrule
5 & Discrete  & \ding{55} & \ding{51} & 1 &1.25 &82.6 &89.6 &1.25 &96.6 &61.2 \\
6 & Discrete  & \ding{51} & \ding{51} & 1 &2.74      & 92.8 & 97.8 &2.74       & 96.8  & 80.8 \\
\midrule
7 & Continuous & \ding{51} & \ding{55} & 0 &24.94      & 80.0 & 89.6 &48.20       & 84.0  & 48.6 \\
8 & Continuous & \ding{51} & \ding{55} & 1 &14.97      & 90.2 & 92.4 &28.30       & 88.4  & 67.0 \\
9 & Continuous & \ding{51} & \ding{51} & 1 &7.75      & 91.4 & 99.4 &15.78       & 98.2  & 81.4 \\
\bottomrule
\end{tabular}
\end{table*}

\begin{figure*}[t]
  \centering
  \includegraphics[width=0.96\linewidth]{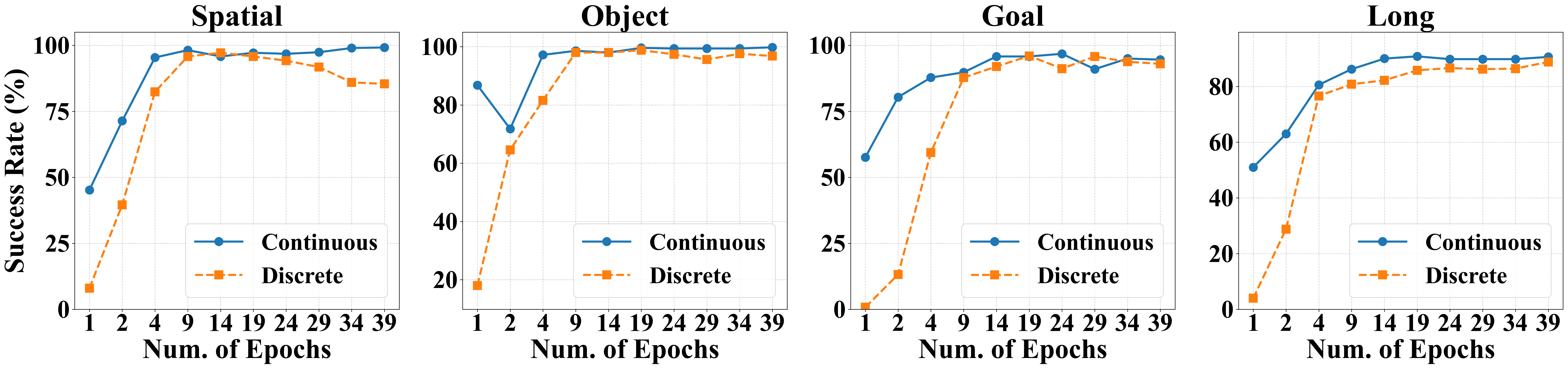}
   \caption{Performance comparison between discrete action and continuous action.}
   \label{fig:action_dis_conti}
\end{figure*}

\noindent \textbf{Discrete Action and Continuous Action.}  
Our model supports both discrete and continuous action generation, and the experimental results reveal a clear advantage for the latter. Concretely, in the LIBERO simulation benchmark, continuous actions result in significantly faster convergence (Fig.~\ref{fig:action_dis_conti}). Although the final performances of these two models in simulation are comparable, their performance gap becomes considerably significant in real-robot experiments. As shown in Table~\ref{tab:ab_real_action}, our real-world experiments demonstrate a much more substantial improvement when using continuous actions over discrete ones.

\begin{wraptable}[7]{r}{0.52\textwidth}
\caption{Ablation study of world model pretraining.
}
\centering
\setlength{\tabcolsep}{1 mm}
\begin{tabular}{lcccccc}
\toprule
 & {Goal} & {Object} & {Spatial} & {Long} \\
\midrule
w/o World Model Pretrain    & 67.3          & 82.9           & 77.8            & 23.0        \\
w/ World Model Pretrain      & \textbf{73.1}  &\textbf{84.0} &\textbf{79.8} &\textbf{30.2}  \\
\bottomrule
\label{tab:pretrain}
\end{tabular}
\end{wraptable}
\noindent \textbf{World Model Pretraining for VLA Model.}  
Our RynnVLA-002 unifies VLA model and world model into a single training stage. We further investigate the possibility of cold starting a VLA model with world model pretraining. This form of pretraining necessitates that the model develop an understanding of visual inputs, actions, and the underlying physical dynamics governing state transitions. We pretrain the model using the same data source as the VLA model. As presented in Table~\ref{tab:pretrain}, employing the world model for pretraining leads to notable improvements in grasping performance. These findings highlight the potential of leveraging world model pretraining in robotic applications, particularly in enhancing task-specific performance through prior exposure to general world knowledge.

\noindent \textbf{Real-World Scope and Cost.}
Our current real-world evaluation focuses on SO100 pick-and-place manipulation, which is sufficient to verify the action-world training signal under controlled clutter and distractor settings but does not yet cover broader robot platforms, long-horizon mobile manipulation, or deformable-object tasks. Scaling these experiments is computationally expensive in the present autoregressive image-token formulation: training a single real-world task currently takes roughly four days because image generation introduces many tokens. We view more robot platforms, more sophisticated manipulation tasks, and more efficient image-token generation as important next steps.

\section{Conclusion}

In this work, we propose RynnVLA-002, a unified framework that integrates the VLA and world model, demonstrating that they enhance each other. Through this contribution, we aim to offer the embodied AI research community a concrete methodology for enabling the synergistic interplay between the VLA and world model. Furthermore, we believe this work helps lay the groundwork for a unified foundation for multi-modal understanding and generation that spans text, vision, and action.

\bibliographystyle{assets/plainnat}
\bibliography{paper}


\end{document}